\documentclass[journal]{IEEEtran}

\usepackage{amsmath,amssymb,amsfonts}
\usepackage{algorithmic}
\usepackage{algorithm}
\usepackage{graphicx}
\usepackage{textcomp}
\usepackage{xcolor}
\usepackage{multirow}
\usepackage{float}
\usepackage{color}
\usepackage{booktabs}
\usepackage{xr-hyper}
\usepackage[hidelinks]{hyperref}

\newcommand{\rev}{}
\newcommand{\revv}{}

\ifCLASSOPTIONcompsoc
    \usepackage[caption=false, font=normalsize, labelfont=sf, textfont=sf]{subfig}
\else
\usepackage[caption=false, font=footnotesize]{subfig}
\fi

\author{
        Yuji Cao, 
        Huan Zhao,~\IEEEmembership{Member,~IEEE,}
        Yuheng Cheng,~\IEEEmembership{Student Member,~IEEE,}
        Ting Shu, 
        Yue Chen, ~\IEEEmembership{Member,~IEEE,}
        Guolong Liu,~\IEEEmembership{Member,~IEEE,}
        Gaoqi Liang,~\IEEEmembership{Member,~IEEE,}
        Junhua Zhao,~\IEEEmembership{Senior Member,~IEEE,}
        Jinyue Yan, 
        Yun Li,~\IEEEmembership{Fellow,~IEEE}
\thanks{\textit{Corresponding author: Huan Zhao, Junhua Zhao.}}
\thanks{Yuji Cao and Yue Chen are with the Department of Mechanical and Automation Engineering, The Chinese University of Hong Kong, Hong Kong SAR, 999077, China (email: yjcao@mae.cuhk.edu.hk, yuechen@mae.cuhk.edu.hk)}
\thanks{Huan Zhao and Jinyue Yan are with the Department of Building Environment and Energy Engineering, The Hong Kong Polytechnic University, Hong Kong, 100872, China (email: huan-paul.zhao@polyu.edu.hk, j-jerry.yan@polyu.edu.hk)}
\thanks{Yuheng Cheng, Guolong Liu, and Junhua Zhao are with the School of Science and Engineering, The Chinese University of Hong Kong, Shenzhen, 518172, China, and also with the Center for Crowd Intelligence, Shenzhen Institute of Artificial Intelligence and Robotics for Society (AIRS), Shenzhen, 518129, China (email: yuhengcheng@link.cuhk.edu.cn, liuguolong@cuhk.edu.cn, zhaojunhua@cuhk.edu.cn)}
\thanks{Ting Shu is with Guangdong-Hongkong-Macao Greater Bay Area Weather Research Center for Monitoring Warning and Forecasting (Shenzhen Institute of Meteorological Innovation), Shenzhen, 518125, China. (email: shuting@gbamwf.com)}
\thanks{Gaoqi Liang is with the School of Mechanical Engineering and Automation,
Harbin Institute of Technology, Shenzhen, 518055, China (e-mail:
lianggaoqi@hit.edu.cn)}
\thanks{Yun Li is with the Shenzhen Institute for Advanced Study, University of Electronic Science and Technology of China, Shenzhen, 518110, China, and also with i4AI Ltd., WC1N 3AX London, U.K (email: Yun.Li@ieee.org)}
}

\begin{document}
\bstctlcite{IEEEexample:BSTcontrol}

\title{Survey on Large Language Model-Enhanced Reinforcement Learning: Concept, Taxonomy, and Methods}

\markboth{IEEE TRANSACTIONS ON NEURAL NETWORKS AND LEARNING SYSTEMS, VOL. X, NO. X, XXX 2024}%
{Shell \MakeLowercase{\textit{et al.}}: Bare Demo of IEEEtran.cls for IEEE Journals}

\maketitle

\begin{abstract}
    With extensive pre-trained knowledge and high-level general capabilities, large language models (LLMs) emerge as a promising avenue to augment reinforcement learning (RL) in aspects such as multi-task learning, sample efficiency, and \rev{high-level }task planning. In this survey, we provide a comprehensive review of the existing literature in \textit{LLM-enhanced RL} and summarize its characteristics compared to conventional RL methods, aiming to clarify the research scope and directions for future studies. Utilizing the classical agent-environment interaction paradigm, we propose a structured taxonomy to systematically categorize LLMs' functionalities in RL, including four roles: information processor, reward designer, decision-maker, and generator. For each role, we summarize the methodologies, analyze the specific RL challenges that are mitigated, and provide insights into future directions. Lastly, comparative analysis of each role, potential applications, prospective opportunities and challenges of the \textit{LLM-enhanced RL} are discussed. \rev{By proposing this taxonomy, we aim to provide a framework for researchers to effectively leverage LLMs in the RL field, potentially accelerating RL applications in complex applications such as robotics, autonomous driving, and energy systems.}
\end{abstract}

\begin{IEEEkeywords}
    Reinforcement learning (RL), large language models (LLM), vision-language models (VLM), multimodal RL, LLM-enhanced RL.
\end{IEEEkeywords}

\section{Introduction}

\IEEEPARstart{R}{einforcement} learning (RL) is a powerful learning paradigm that focuses on control and decision-making, where an agent learns to optimize a specified target through trial and error interactions with the environment. \rev{Traditional RL methods, however, often struggled with high-dimensional state spaces and complex environments~\cite{sutton2018reinforcement}. The integration of deep learning techniques with RL, known as deep RL, has led to significant breakthroughs. In 2015, Deep Q-Networks (DQN)~\cite{mnih2015human} marked a turning point, demonstrating human-level performance on Atari games using raw pixel inputs. Subsequent innovations such as proximal policy optimization~\cite{schulman2017proximalRev} and soft actor-critic~\cite{haarnoja2018softRev} have further expanded the capabilities of deep RL. In different realms, deep RL algorithms have achieved promising performance, such as real-time strategy games~\cite{vinyals2019grandmaster},~\cite{berner2019dota}, board games~\cite{silver2016mastering},~\cite{schrittwieser2020mastering}, energy management~\cite{zhao2022mobileRev} and imperfect information games~\cite{schmid2021re},~\cite{brown2020superhuman}.} \rev{Concurrent advancements in natural language processing (NLP) and computer vision (CV)~\cite{vaswani2017attention,krizhevsky2012imagenet}, have fostered new RL paradigms, such as language-conditional RL~\cite{jiang2019language}, which uses natural language to instruct agents, and vision-based RL~\cite{anderson2018vision}, where agents learn from high-dimensional visual inputs.}

\rev{The integration of language and vision capabilities into deep RL has introduced new challenges, as the agent has to learn the high-dimensional features and a control policy jointly.} To reduce the burden of visual feature learning, reference~\cite{stooke2021decoupling} decoupled representation learning from RL. To handle language-involved tasks, a survey~\cite{luketinaSurveyReinforcementLearning2019} called for potential uses of NLP techniques in RL. Nevertheless, the capabilities of language models were limited at that time and the following four challenges still have not been addressed: \rev{1) \textit{Sample inefficiency}: Language and vision tasks involve large, complex state-action spaces, making it challenging for RL agents to learn effective policies. Moreover, agents must understand tasks and connect them to corresponding states, necessitating even more extensive interactions~\cite{mandlekar2021matters},~\cite{lynch2022interactive},~\cite{Yang2022TowardsAR}. \rev{2) \textit{Reward function design}: In language and vision tasks, designing effective reward functions is particularly challenging. These functions must capture subtle linguistic nuances and complex visual features, significantly increasing the complexity of an already difficult process. Moreover, aligning rewards with high-level task objectives in these domains often requires domain expertise and extensive trial-and-error~\cite{knox2023reward},~\cite{dworschak2022reinforcement},~\cite{sutton2018reinforcement},~\cite{booth2023perils}.} 3) \textit{Generalization}: RL agents often overfit to training data, especially in vision-based environments, leading to poor performance when deployed in states with interventions (e.g., added noise). Agents must learn invariant features robust to such interventions, enabling generalization across varied linguistic contexts and visual scenes. However, the complexity of these domains makes extracting such features and adapting to new environments particularly challenging~\cite{di2022goal},~\cite{yang2019generalized}.} 4) \textit{Natural language understanding}: deep RL faces difficulties in natural language processing and understanding scenarios, where the nuances and complexities of human language present unique challenges that are not adequately addressed by current RL methodologies~\cite{luketina2019survey}.

\rev{The field of NLP has undergone a revolutionary transformation since the introduction of the Transformer architecture in 2017~\cite{vaswani2017attention}. This breakthrough paved the way for the development of Large Language Models (LLMs), with landmark models such as BERT~\cite{devlin2018bert}, GPT~\cite{radford2018improving}, and more recent iterations \revv{such as} GPT-3~\cite{brown2020language} and PaLM~\cite{chowdhery2023palm} marking significant milestones. The emergence of these LLMs has demonstrated powerful capabilities across various real-world applications, including medicine~\cite{thirunavukarasu2023large}, chemical research~\cite{boiko2023autonomous}, energy system~\cite{jiang2024eplusRev},~\cite{tan2024generalRev},~\cite{zhou2024elecbenchRev} and embodied control in robotics~\cite{liang2023code}. These models have not only advanced the field of NLP but also shown remarkable potential in tackling complex, multi-disciplinary challenges.} Compared to small language models, LLMs have emergent capabilities that are not present in small language models~\cite{webb2023emergent}, such as in-context learning~\cite{wei2023larger}, reasoning ability~\cite{huang2022towards} etc. Additionally, leveraging the vast amounts of training data, pre-trained LLMs are equipped with a broad spectrum of world knowledge~\cite{yu2023kola}. Benefiting from these capabilities, the applications of language models have been shifted from language modeling to task-solving, ranging from basic text classification and sentiment analysis to complex high-level task planning~\cite{singh2023progprompt} and decision-making~\cite{stiennon2022learning},~\cite{akyürek2023learning}. 

With emergent capabilities, the potential of LLMs to address the inherent challenges of RL has recently gained popularity~\cite{duGuidingPretrainingReinforcement2023},~\cite{cartaGroundingLargeLanguage2023}. The capabilities of LLMs, particularly in natural language understanding, reasoning, and task planning, provide a unique approach to solving the above-mentioned RL issues. For sample inefficiency, reference~\cite{linLearningModelWorld2023} proposed a framework where LLMs can be employed to improve the sample efficiency of RL agents by providing rich, contextually informed predictions or suggestions, thereby reducing the need for extensive environment interactions. For reward function design, LLMs can aid in constructing more nuanced and effective reward functions, enhancing the learning process by offering a deeper understanding of complex scenarios~\cite{li2023auto}. For generalization, reference~\cite{chakrabortyREMOVEAdaptivePolicy2023} proposed a framework that leverages language-based feedback for improving the generalization of RL policy in unseen environments. For natural language understanding, Pang \textit{et al.} ~\cite{pang2023naturallanguageconditionedreinforcementlearning} used LLMs to translate complex natural language-based instructions to simple task-specified languages for RL agents. These works show that LLM is a promising and powerful role that can contribute to the longstanding RL challenges. 

Despite the advancements in the domain of integrating LLMs into the RL paradigm, there is currently a notable absence of comprehensive review in this rapidly evolving area. Additionally, though various methods are proposed to integrate LLMs into the RL paradigm, there is no unified framework for such integration. Our survey paper seeks to fill these gaps by providing an extensive review of the related literature, defining the scope of the novel paradigm called \textit{LLM-enhanced RL}, and further proposing a taxonomy to categorize the functionalities of LLMs in the proposed paradigm.

\subsection{Contributions}
This survey makes the following contributions:

\begin{itemize}
    \item \textit{LLM-enhanced RL paradigm}: This paper presents the first comprehensive review in the emerging field of integrating LLM into the RL paradigm. To clarify the research scope and direction for future works, we define the term \textit{LLM-enhanced RL} to encapsulate this class of methodologies, summarize the characteristics and provide a corresponding framework that clearly illustrates 1) how to integrate LLMs in classical agent-environment interaction and 2) the multifaceted enhancements that LLMs offer to the conventional RL paradigm.
    \item \textit{Unified taxonomy}: Further classifying the functionalities of LLMs in the LLM-enhanced RL paradigm, we propose a structured taxonomy to systematically categorize LLMs within the classical agent-environment paradigm, where LLMs are classified as information processors, reward designers, decision-makers, and generators. By such a categorization, a clear view of how LLMs integrate into the classical RL paradigm is offered.
    \item \textit{Algorithmic review}: For each role of LLM, we review emerging works in this direction and discuss different algorithmic characteristics from the perspective of capabilities. Based on this foundation, future applications, opportunities, and challenges of LLM-enhanced RL are analyzed to provide a potential roadmap for advancing this interdisciplinary field.
\end{itemize}
\subsection{Text Organization}
The remaining sections are organized as follows. Section~\ref{sec:background} provides foundational knowledge of both RL and LLM. Section~\ref{sec:framework} presents the concept of LLM-enhanced RL and provides its overall framework. Following this, Sections~\ref{sec:information-processor},~\ref{sec:reward-designer},~\ref{sec:decision-maker}, and~\ref{sec:generator} offer an in-depth analysis of LLMs within the RL context, exploring their roles as information processor, reward designer, decision-maker, and generator, respectively. Last, Section~\ref{sec:discussion} discusses the application, opportunities and challenges of LLM-enhanced RL. Finally, Section~\ref{sec:conclusion} concludes the survey.

\section{Background}
\label{sec:background}
In this section, we provide a concise overview of the classical RL paradigm and related challenges. Next, we explore the prevailing trend in RL—specifically, the fusion of multimodal data sources, including language and visual information. Following this, we offer an introductory background on LLMs and outline the key capabilities that can enhance the RL learning paradigm.
\subsection{Background of Reinforcement Learning}
\subsubsection{Classical Reinforcement Learning}
\begin{figure}[bthp]
    \centering
    \includegraphics[width=0.95\linewidth]{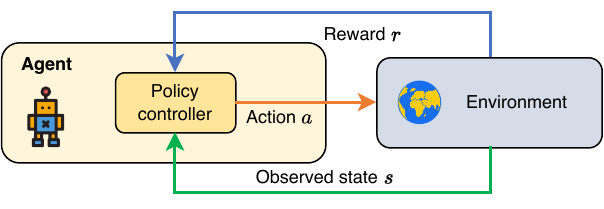}
    \caption{Classical reinforcement learning paradigm.}
    \label{fig:classical-RL}
\end{figure}

In a classical RL paradigm shown in Fig.~\ref{fig:classical-RL}, the agent interacts with an environment through a trial-and-error process to maximize the specified rewards in the trajectory. In each step, the agent takes an action $a$ based on the observed state $s$ from the environment. By optimizing the policy $\pi$ (action controller), the agent maximizes the cumulative rewards. Such an optimization problem is usually formalized through the concept of Markov Decision Process (MDP), defined by the quintuple \( \langle \mathcal{S}, \mathcal{A}, \mathcal{T}, \mathcal{R}, \gamma \rangle \). Here, \( \mathcal{S} \) denotes a set comprising all possible states, \( \mathcal{A} \) denotes a set of all possible actions, \( \mathcal{T} \) represents the state transition probability function \( \mathcal{T}: \mathcal{S} \times \mathcal{A} \times \mathcal{S} \rightarrow [0,1] \), \( \mathcal{R} \) is a reward function \( \mathcal{R}: \mathcal{S} \times \mathcal{A} \times \mathcal{S} \rightarrow \mathbb{R} \), and \( \gamma \) (with \( 0 \leq \gamma \leq 1 \)) is the discount factor. The objective in RL is to optimize the policy \( \pi(a|s) \) such that the cumulative returns \( \sum_{k=0}^{\infty} \gamma^k r_{k+1} \) is maximized.

\subsubsection{Challenges of Reinforcement Learning} While RL algorithms have made remarkable performance in recent years~\cite{mnih2015human},~\cite{silver2016mastering},~\cite{kalashnikov2018qtopt}, there are still several longstanding challenges that limit the real-world applicability of RL:
\begin{itemize}
    \item \textit{Generalization in Unseen Environment}: Generalization in unseen environments remains a significant challenge in the field of RL~\cite{wang2020improving}. The core issue lies in the ability of RL algorithms to transfer learned knowledge or behaviors to new, previously unseen environments. RL models are often trained in simulated or specific settings, excelling in those scenarios but struggling to maintain performance when faced with novel or dynamic conditions. This limitation hinders the practical application of RL in real-world situations, where environments are rarely static or perfectly predictable. Achieving generalization requires models to not only learn specific task solutions but also to understand underlying principles that can be adapted to a range of situations.
    \item \textit{Reward Function Design}: Reward function is the principal contributing factor to the performance of an RL agent. Despite their fundamental importance, reward functions are known to be notoriously difficult to design, especially in contexts involving sparse reward environments and complex scenarios~\cite{sutton2018reinforcement}. In sparse reward settings, where feedback is limited, reward shaping becomes essential to guide agents toward meaningful behaviors; however, this introduces the risk of inadvertently biasing the agent towards sub-optimal policies or overfitting to specific scenarios~\cite{devidze2022exploration},~\cite{singh2019end}. Conversely, for complex tasks, high-performance reward functions usually require massive manual trial-and-error since most designed rewards are sub-optimal~\cite{booth2023perils} or lead to unintended behavior~\cite{hadfield2017inverse}.
    \item \textit{Compounding Error in Model-based Planning}: Model-based RL is prone to the issue of compounding errors during planning. As the prediction horizon extends, errors in the model's predictions accumulate, leading to significant deviations from optimal trajectories~\cite{xiao2019learning},~\cite{moerland2023model}. This problem is particularly acute in complex environments with high-dimensional state spaces. With their advanced predictive capabilities and understanding of sequential dependencies, LLMs could help mitigate these errors, leading to more accurate and reliable planning in model-based RL.
    \item \rev{\textit{Multi-task Learning:} Multi-task RL faces several key challenges that limit its effectiveness. One major issue is managing varying task difficulties, where simpler tasks can overshadow learning of more complex ones, leading to negative transfer~\cite{cho2024hardRev}. Task interference is another critical problem, as shared parameters or data between tasks can result in suboptimal performance on individual tasks~\cite{feng2024efficientmultitaskreinforcementlearningRev}. Determining optimal parameter-sharing strategies is complex, as it must balance learning efficiency with task-specific requirements~\cite{sun2022pacoRev}. Sample efficiency remains a significant hurdle, with traditional data-sharing approaches not fully leveraging learned behaviors across tasks~\cite{zhang2024efficientRev}. Finally, effectively transferring knowledge between tasks without negative interference is an ongoing challenge that impacts the agent's ability to accelerate learning across multiple objectives~\cite{electronics9091363Rev}.}
\end{itemize}
\subsubsection{Multimodal Reinforcement Learning}
With the advances in both CV and NLP, pattern recognition in vision and natural language has become increasingly powerful, and multimodal data has been involved in the RL paradigm recently. Visual data is commonly involved in the observation space of RL when agents receive image-based information from the environment, e.g. in applications such as robots~\cite{xiaoRoboticSkillAcquisition2023}, video game control~\cite{mnih2015human} etc. Compared to visual data, natural languages are usually included when RL agents are given specific tasks when interacting with the environments. The use of natural languages in RL can be divided into the following two categories~\cite{luketinaSurveyReinforcementLearning2019}:
\begin{itemize}
    \item \textit{Language-conditional RL}: In language-conditional RL, the problem itself requires the agent to interact with the environment through language. Specifically, there are two ways to integrate natural language in RL: 1) \textit{task description}: the task or instruction is described in natural languages, \textit{e.g.} instruction following, where the agents learn to interpret the instructions first and then execute actions; 2) \textit{action space or observation space}: natural language is part of the state and action space, \textit{e.g.}. text games, dialogue systems, and question answering (Q\&A). This class of RL leverages natural language as a direct component of the RL process, guiding the agent's actions and decisions within the language environment.
    \item \textit{Language-assisted RL}: In language-assisted RL, natural language is used to facilitate learning but not as a part of problem formulation. Two usages of language-assisted RL are: 1) \textit{communicating domain knowledge}: the text containing task-related information can be helpful for agents. Therefore, wikis and manuals related to the environment can potentially assist the agents in such cases; 2) \textit{structuring policies}: structuring policies is to communicate information about the state or dynamics of the environment based on language instead of representations of the environment or models. In such cases, language can be leveraged to shape representations towards a generalizable abstraction, such as using ``avoid hitting the wall'' instead of representations of a policy. This approach represents a more indirect use of natural language, serving as a guide or enhancer to the primary RL tasks.
\end{itemize}

The integration of multimodal data challenges the RL paradigm since the agent has to simultaneously learn how to process complex multimodal data and optimize the control policy in the environment~\cite{stooke2021decoupling}. Issues such as natural language understanding~\cite{yuan2023plan4mc} and visual-based reward function design~\cite{rocamondeVisionLanguageModelsAre2023} require to be addressed.
\subsection{Background of Large Language Models}

LLMs typically refer to the Transformer-based language models~\cite{Vaswani2017AttentionIA} containing billions of parameters and being trained on massive text data (i.e., several terabytes (TB) scale)~\cite{Shanahan2022TalkingAL32}, such as GPT-3~\cite{brown2020language} and LLaMA~\cite{Touvron2023LLaMAOA57}. \rev{The extensive number of parameters and internet-scale training data enable LLMs to master a diverse array of tasks, resulting in enhanced capabilities in language generation, knowledge representation, and logical reasoning, as well as improved generalization to novel tasks.}

The development and effectiveness of LLMs are largely driven by \textit{Scaling Laws}, i.e., as these models grow in size – both in terms of their parameter count and the data they are trained on – they tend to exhibit \textit{emergent abilities} that are not present in small models~\cite{Kaplan2020ScalingLF},~\cite{Hoffmann2022TrainingCL},~\cite{Wei2022EmergentAO}, such as in-context learning, reasoning, and generalization. Here, we briefly introduce such capabilities of LLMs in detail:

\begin{itemize}
    \item \textit{In-context Learning}: In-context learning capability eliminates the need for explicit model retraining or gradient update~\cite{brown2020language}, as it can generate better responses or perform tasks by inputs cueing examples or related knowledge. Specifically, task-related texts are included in the prompts as context information, helping the LLMs to understand the situations and execute instructions.
    \item \textit{Instruction Following}: Leveraging diverse task-specific datasets formatted with natural language descriptions (also called \textit{instruction tuning}), LLMs are shown to perform well on unseen tasks that are also described in the form of natural language~\cite{sanh2021multitask},~\cite{ouyang2022training},~\cite{cheng2024gaia}. Therefore, this capability equips LLMs with the ability to comprehend instructions for new tasks and effectively generalize across tasks not previously encountered, even in the absence of explicit examples.
    \item \textit{Step-by-step Reasoning}: For smaller models, tackling multi-step tasks, such as solving math word problems, often proves to be challenging. However, large language models can address the complex task effectively with sophisticated prompting strategies such as Chain of Thought (CoT)~\cite{wei2023chainofthought}, Tree of Thought (ToT)~\cite{yao2023tree}, and Graph of Thought (GoT)~\cite{besta2023graph}. These strategies structure the problem-solving process into sequential or hierarchical steps, facilitating a more articulated and understandable reasoning pathway. Additionally, prompts designed for planning enable LLMs to output sequences that reflect a progression of thoughts or actions, proving invaluable for tasks demanding logical sequence or decision-making outputs.
\end{itemize}

\section{Large Language Model-Enhanced Reinforcement Learning}
\label{sec:framework}

\begin{figure*}[htbp]
    \centering
    \includegraphics[width=\linewidth]{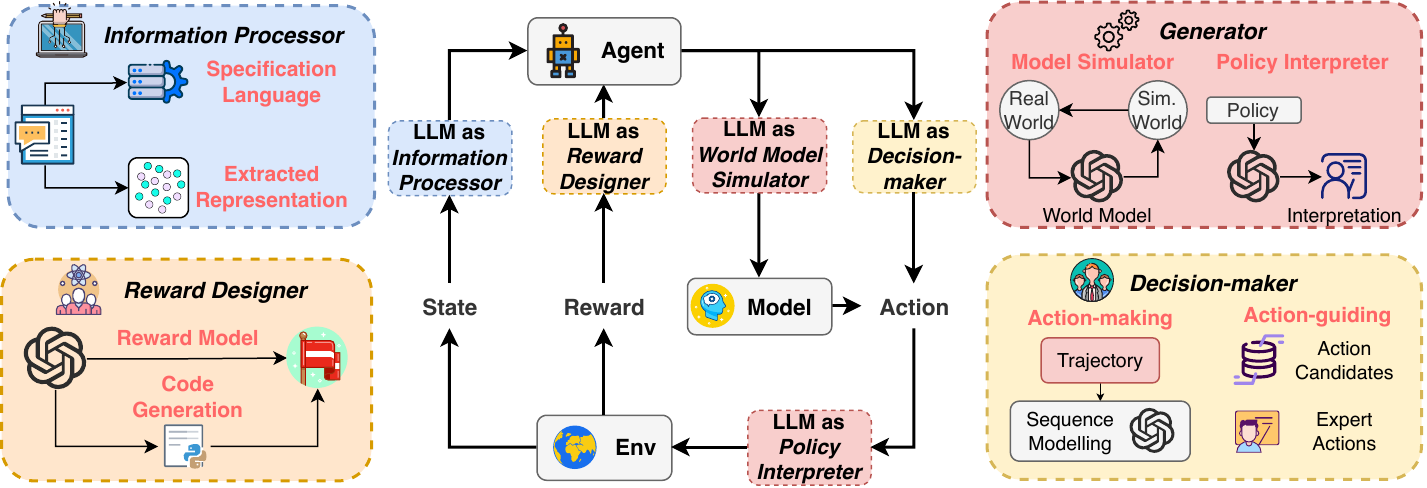}
    \caption{Framework of LLM-enhanced RL in classical Agent-Environment interactions, where LLM plays different roles in enhancing RL.}
    \label{fig:framework}
\end{figure*}

\subsection{Definition}
RL agents are often tasked with making robust and deliberate decisions using multimodal information in real-real applications, whether in the MDP setting or within the context of specific task descriptions. Examples include robots designed to follow natural language instructions while navigating physical environments or visual games with tasks described in natural language~\cite{rao2021visual},~\cite{huang2021deductive},~\cite{yang2020multitask}. However, it is challenging for conventional RL methods as the agent is required to simultaneously interpret complex multimodal data and optimize control policies amidst ever-changing environments~\cite{he2023derl}. Compounding these challenges are issues like sample inefficiency, the difficulty of crafting reward functions that accurately reflect multimodal inputs, and the need for robust generalization across varied tasks and settings.

The rapid advancements in LLMs present a viable solution to these challenges, thanks to their potent natural language understanding and reasoning abilities, coupled with recent progress in incorporating visual data processing~\cite{liu2023survey}. This dual capability enables LLMs to interpret and act upon complex multimodal information effectively, serving as a robust helper for enhancing the RL paradigm for real-world applications. 

Nevertheless, despite the powerful functionalities of LLMs, the current studies are varied and lack a standard concept correctly specifying the systematic methodology, which impedes the advancement of research in this area. Therefore, we introduce the concept called \textit{LLM-enhanced RL} as follows:

\rev{\textit{\textbf{LLM-enhanced RL} refers to the methods that utilize the multimodal information processing, generating, reasoning, and other high-level cognitive capabilities of pre-trained, knowledge-inherent LLM models to assist the RL paradigm.}}

\rev{LLM-enhanced RL differs from traditional model-based RL by leveraging knowledge-rich LLM models. This approach provides two key advantages: First, LLM equips the agent with substantial pre-trained capabilities at the beginning of the learning process, such as reasoning and high-level planning, etc. Second, it offers superior generalization capabilities. Pre-trained on diverse data, LLMs can effectively transfer knowledge across domains, enabling better adaptation to unseen environments than conventional data-driven models. Last, LLM-enhanced RL addresses a key limitation of pre-trained models: their inability to interact with environments to expand their knowledge and ground themselves in specific domains. Through environmental interactions, this approach generates task-specific data, grounds the LLM in particular domains by in-context learning, and eventually helps them adapt to dynamic changes with continuous learning.}

\subsection{Framework}
The framework of LLM-enhanced RL is illustrated in the center of Fig.~\ref{fig:framework}, which is founded on the classical agent-environment interaction paradigm. Along with the trial-and-error learning process, LLM processes the state information, redesigns the reward, assists in action selection, and interprets the policy after the action selection.

Specifically, on the one hand, when the agent receives the state and reward information from the environment, LLM is able to process or modify the information to either filter unnecessary natural language-based information or design appropriate rewards to accelerate the learning process, based on the natural language understanding and reasoning capabilities. On the other hand, when the agent is about to choose an action based on the observation, LLM can assist the action selection process by either simulating a world model or serving as the policy network to generate reasonable actions based on the modeling capability and common-sense knowledge. Additionally, after the action selection process, integrating state, reward, and action information, LLM can interpret the underlying possible reasons behind the policy selection, which helps human supervisors understand the scenarios for further system optimization.

Based on the functions of LLM in the framework, we extract characteristics of LLM-enhanced RL and further divide four different LLM roles in LLM-enhanced RL, including information processor, reward designer, generator, and decision-maker, which will be elaborated in the next subsections.

\subsection{Characteristics}

The LLM-enhanced RL paradigm enhances the vanilla RL paradigm with the following characteristics:
\begin{itemize}
\item \textbf{Multimodal Information Understanding}: LLMs enhance RL agents' comprehension of scenarios involving multimodal information, enabling them to learn from tasks or environments described in natural language and vision data more effectively.
\item \textbf{Multi-task Learning and Generalization}: Benefiting from \rev{the multi-disciplinary pre-trained knowledge and powerful sequence modeling capability, LLMs empower RL agents by providing a high-capacity model capable of accommodating task variances and transferring knowledge across multiple tasks, assisting in handling multiple tasks}. 
\item \textbf{Improved Sample Efficiency}: Given the inherent exploratory nature, the RL paradigm demands significant samples to learn. Pre-trained LLM can enhance data generation by simulation or leverage the prior knowledge to improve the sample efficiency of RL.
\item \textbf{Long-Horizon Handling}: RL becomes more challenging as the length of trajectory increases, due to the credit assignment problem. LLMs can decompose complex tasks down into sub-tasks to assist RL agents in planning over longer temporal horizons, aiding in the decision-making process for complex, multi-step tasks such as the Minecraft game.
\item \textbf{Reward Signal Generation}: Based on the context understanding and domain knowledge, LLMs contribute to the reward shaping and reward function designing, which help guide the RL towards effective policy learning in sparse-reward environments.
\end{itemize}

\subsection{Taxonomy}
In this subsection, we illustrate the different roles of LLMs within the above framework, by detailing their functions and corresponding issues of RL they address:

\begin{itemize}
    \item \textbf{Information Processor}: When observation or task description involves language or visual features, it is challenging for the agent to comprehend the complex information and optimize the control policy simultaneously. To release the agent from the burden of understanding the multimodal data, LLM can serve as an information processor for environment information or task instruction information by 1): extracting meaningful feature representations for speeding up network learning; 2) translating natural language-based environment information or task instruction information into formal specific task languages to reduce learning complexity. 
    
    \rev{\textit{Application example}: In instruction-following RL for robots, tasks can have unbounded natural language forms due to users' diverse speaking habits, which can impede RL learning performance. LLMs transform these varied natural language instructions into a unique task language, enabling more robust RL performance~\cite{pang2023naturallanguageconditionedreinforcementlearning}.}
    \item \textbf{Reward Designer}: In complex task environments where the reward is sparse or a high-performance reward function is hard to define, using the prior world knowledge, reasoning abilities, and code generation ability, LLM can serve as two roles: 1) an implicit reward model to provide reward values based on the environment information, either by training or prompting; 2) an explicit reward model that generates executable codes of reward functions that transparently specifies the logical calculation process of reward scalars based on the environment specifications and language-based instructions or goals. 
    
    \rev{\textit{Application example}: For complex robotic control problems, such as dexterous manipulation, reward design requires both expertise and trial and error. LLM designs reward functions based on knowledge and iteratively improves it based on the performance~\cite{maEurekaHumanLevelReward2023}.}
    \item \textbf{Decision-maker}: \rev{RL faces challenges such as sample inefficiency and exploration inefficiency. To address these challenges, LLMs can be leveraged as decision-makers in RL, offering promising solutions through two main approaches: 1) action-making: LLMs treat offline RL as a sequence modeling problem, using rewards for the conditional generation of actions. Pretrained on diverse, internet-scale data, LLMs possess advanced semantic understanding capabilities, which can be exploited to accelerate offline RL learning. 2) action-guiding: LLMs act as expert instructors to produce a reduced set of action candidates or expert actions. The action candidates constrain the original action space, thereby enhancing exploration efficiency. Expert actions encapsulate prior knowledge from LLMs. When incorporated to regularize the policy learning, this expert knowledge is distilled into an RL agent, resulting in better sample efficiency.}
    
    \rev{\textit{Application example}: In embodied robot, given human-instruction and language-based description, LLM generates potential actions for the robot to choose from~\cite{ahnCanNotSay2022}.}
    \item \textbf{Generator}: Model-based RL hinges on precise world models to learn accurate environment dynamics and simulate high-fidelity trajectories. Additionally, interpretability remains another important issue in RL. Using the multimodal information understanding capability and prior common-sense reasoning ability, LLMs can be 1) a generator to generate accurate trajectories in model-based RL; 2) generate policy explanations with the prompts of related information in explainable RL.
    \rev{\textit{Application example}: In Minecraft item crafting, LLMs generate Abstract World Models---hypothesized sequences of subgoals for a given task. These LLM-generated world models guide RL agents' exploration and learning and the RL agent verifies and corrects the world model through gameplay, combining LLM knowledge with grounded experience to achieve an order of magnitude improvement in sample efficiency over traditional methods~\cite{nottinghamEmbodiedAgentsDream2023}.}
\end{itemize}

\section{LLM as Information Processor}
\label{sec:information-processor}
The normal way for deep RL with language or visual information is to jointly process the information and learn a control policy, end-to-end. However, this demands the RL agent to learn to comprehend the information and manage the task simultaneously. Additionally, learning the language or visual features by simply relying on the reward function is challenging and may narrow the learned features to a narrow utility, hampering the generalization ability of the agent~\cite{stooke2021decoupling}. 

With the advances in unsupervised techniques and large-scale pre-trained models for CV and NLP, the decoupled structure, where the encoders are separately trained, has gained popularity~\cite{stooke2021decoupling},~\cite{srinivasCURLContrastiveUnsupervised2020},~\cite{schwarzer2020data}. Utilizing the powerful representation ability and prior knowledge, the pre-trained LLM or vision-language model (VLM) model can serve as an information processor for RL. \rev{They can extract observation representations for downstream networks or translate natural language into formal specifications, enabling the execution of multiple tasks. This multi-task capability improves sample efficiency and zero-shot performance, allowing agents to generalize effectively across diverse and sparse-reward environments.}
\subsection{Feature Representation Extractor}
\label{subsec:feature-representation-extractor}

\begin{figure}[htbp]
    \centering
    \includegraphics[width=\linewidth]{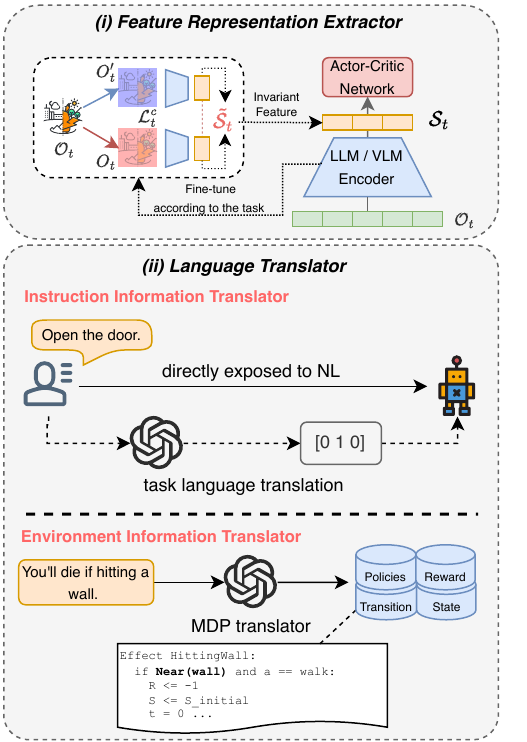}
    \caption{LLM as an information processor. \rev{(i) Feature Representation Extractor: frozen/fine-tuned LLM extracts meaningful representations for downstream RL networks. In the fine-tuning process, given observation ($\mathcal{O}_t$), invariant feature abstraction ($\Tilde{S}_t$) is learned with the contrastive loss ($\mathcal{L}^{c}_t$). Then, the invariant is fed into the actor-critic network. After fine-tuning, given different observations ($O_t$) and ($O'_t$) with appearance variation, the extracted representation is invariant, leading to robust RL performance. (ii) Language Translator: LLM interprets diverse natural language inputs, converting them into a standardized, task-specific format that the RL agent can efficiently process and act upon.}}
    \label{fig:information-processor}
\end{figure}

Adopting the large pre-trained models in CV and NLP, the learned feature representation can be a scaffold embedding for downstream network learning and increase the sample efficiency. As illustrated in Fig.~\ref{fig:information-processor} (i), the usages can be further divided into two categories according to whether the model is trained simultaneously. One way is to directly use the frozen pre-trained model to extract embeddings from the observation $\mathcal{O}_t$ and another way is to further fine-tune the pre-trained model using contrastive learning with a contrastive loss $\mathcal{L}_t^c$ to achieve better adaptation in new environments.

\subsubsection{Frozen Pre-trained Model}
\label{subsubsec:frozen}
Using the frozen large-scale pre-trained model is the straightforward way. In reference~\cite{paischerHistoryCompressionLanguage2023}, the author proposed History Compression via Language Models (HELM) to utilize a frozen pre-trained Language Transformer to extract history representation and compression and thus addresses the problem of partially observed MDP by approximating the underlying MDP with the past representation. Specifically, the framework first uses FrozenHopfield, a frozen associative memory to map observations $[o_{t-2},o_{t-1},o_t]$ to a compressed representation $h_t$ and then concatenate it with a learned encoding of the current observation via a convolutional neural network. Such a method solves the problem of how to effectively utilize the compressed history for policy optimization. After that, Semantic HELM~\cite{paischerSemanticHELMHumanReadable} proposed a human-readable memory mechanism that summarizes past visual observations in human language and uses multimodal models to associate visual inputs with language tokens. The memory is a semantic database $\mathcal{S}$ constructed by encoding prompt-augmented tokens from the vocabularies of Contrastive Language-Image Pre-training (CLIP)~\cite{radford2021learning} and the pre-trained language models. Given an observation $o_t$, the agent retrieves top-$k$ embeddings from $\mathcal{S}$ as actor-critic input to assist the policy optimization. Such a memory mechanism provides a human-readable representation of the past and helps the agent to cope with partially observable environments. \revv{In the experiment, they used the proximal policy optimization (PPO)~\cite{schulman2017proximalRev} algorithm and pretrained Transformer XL~\cite{dai2019transformerxlattentivelanguagemodels} model. When testing on the partially observable environments, they found the extracted semantics help the memory-less agent obtain comparable scores with memory-based methods trained on long trajectories.\label{rev:helm}} However, one limitation of such frozen pre-trained models is that the representations cannot dynamically adjust according to the task and environment.

\subsubsection{Fine-tuning Pre-trained Model}
\rev{When trained RL agents are deployed in real-world applications, their performance often deteriorates under significant appearance variations (out-of-distribution data) due to overfitting to training scenarios. For example, robots with vision-based navigation tasks may fail when the environment color changes. Invariant feature representations serve as a form of state abstraction that remains consistent across out-of-distribution appearance variations such as added noise, brightness changes, or slight rotations. When encountering appearance changes or out-of-distribution data, though the observation changed, the representation (feature embedding) that fed into the policy/value network is nearly unchanged, leading to robust RL performance and increased generalization in unseen environments.}

\rev{Contrastive learning is a common way to learn the invariant feature representation. It learns representations from high-dimensional data by contrasting positive examples against negatives.} Given a query \( q \), the goal is to match query \( q \) more closely to a positive key \( k_+ \) than to any negative keys \( \mathbb{K} \setminus \{ k_+ \} \) in a set \( \mathbb{K} \). This process is modeled using similarity measures, such as the dot product \( (q^T k) \) or the bilinear product \( (q^T W k) \), where \( W \) is a weight matrix. To effectively learn these representations, a contrastive loss function such as InfoNCE~\cite{oord2018representation} is used:
\begin{equation}\label{eq:contrastive-loss}
    \mathcal{L}^c = \log\frac{\exp(q^TWk_+)}{\exp(q^TWk_+) + \sum_k^{K-1}\exp(q^TWk_i)}
\end{equation}
where $\exp$ is the exponential symbol and the whole $\mathcal{L}^c$ can be viewed as the log-loss of a softmax classifier, treating the matching of \( q \) to \( k_+ \) as a multi-class classification problem among \( K \) classes. By maximizing alignment between different changes of the same observation via the above loss, the model can learn the invariant representations.

\rev{When combining with RL, given different RL tasks, the required invariant feature representations should be adjusted accordingly. Therefore, researchers have explored different ways to improve contrastive learning.} In reference~\cite{choiEfficientPolicyAdaptation2023}, to achieve the zero-shot capability of embodied agents, the author devised a visual prompt-based contrastive learning framework that uses a pre-trained VLM to learn the visual state representations. \rev{The visual prompts are learned on expert demonstrations from domain factors such as camera settings and stride length.} By contrastively training the VLM on a pool of visual prompts along with the RL policy learning process, the learned representations are robust to the variations of environments, \rev{leading an increase of 18-20\% success rates on unseen scenarios and improved generalization capability.} Based on the contrastive learning, another method ReCoRe~\cite{poudelReCoReRegularizedContrastive2023} added an intervention-invariant regularizer in the form of an auxiliary task such as depth prediction and image denoising to explicitly enforce invariance of learned representations to environment changes. 

\subsection{Language Translator}

The unbounded and diverse representation of natural languages in both human instruction and environmental information impedes policy learning. LLM can be leveraged as a language translator to reduce the additional burden of comprehending natural language for RL agents and increase sample efficiency. As illustrated in Fig.~\ref{fig:information-processor} (ii), LLM transforms the diverse and informal natural language information into formal task-specific information, such as feature representation or task-specific languages, thus assisting the learning process of the RL agent.

\subsubsection{Instruction Information Translation}
One application of LLM is to translate natural language-based instructions for instruction-following applications. In reference~\cite{pang2023naturallanguageconditionedreinforcementlearning}, the author investigated an \textit{inside-out} scheme for natural language-conditioned RL by training an LLM that translates the natural language to a task-related unique language. Such an inside-out scheme prevents the policy from being directly exposed to natural language instructions and helps efficient policy learning. Another literature \textit{STARLING}~\cite{STARLINGSelfsupervisedTraining2023} used LLM to translate natural language-based instructions to game information and example game metadata. The translated information is then fed forward to \textit{Inform7}, an interactive fiction game engine, to develop a large amount of text-based games for RL agents to master the desired skills.

\subsubsection{Environment Information Translation}
On the other hand, LLM can also be used to translate the natural language environment information into formal domain-specific language that can specify MDP information, which converts natural language sentences into grounded usable knowledge for the agent. Previous works generally ground natural language into individual task components such as task objectives description~\cite{patel2020grounding}, rewards~\cite{sumers2021learning} and polices~\cite{liangCodePoliciesLanguage2023a},~\cite{song2023llm}. To unify the information about all the components of a task,~\cite{InformingReinforcementLearning2023} introduces \textit{RLang}, a grounded formal language capable of expressing information about every element of an MDP and solutions such as policies, plans, reward functions and transition functions. By using LLM to translate natural language to RLang and train RL agents upon RLang, the agents are capable of leveraging information and avoid having to learn \textit{tabula rasa}. 

\subsection{Summarization and Outlook}
\label{subsec:summarization-info-processor}
For information processing, LLM is used to accelerate RL learning processing by decoupling the information processing task and the controlling task, where LLM extracts feature representations or handles the natural language-based information. 

\label{revv:multimodal}
\revv{When multimodal data are involved in the environment, e.g., robot manipulation tasks, the information processing task becomes more challenging. For one thing, the misalignment or contradiction between different modalities may exist~\cite{Rev2gordon2025mismatch}. Modality weighting techniques that adaptively learn the importance of different modalities by attention mechanisms provide a potential solution for this problem~\cite{Rev2wang2024multimodal}. In addition, CLIP-based multimodal foundation models using large-scale image-text pairs have shown outstanding zero-shot ability in various multimodal tasks~\cite{radford2021learning}. By learning cross-modal and task semantics, the misaligned modality can be replaced with a virtually generated modality~\cite{Rev2zhao2024toward}. For another, how to effectively combine the information between different modalities into a unified representation, i.e., multimodal fusion, is also an important issue. In reference~\cite{Rev2lygerakis2024m2curl}, a multimodal contrastive learning with attention mechanisms, which learns the intra and inter-modal representations, was proposed. The different attention heads align the agreement from one modality to another and vice versa, providing an informative representation for the RL agent. However, most multimodal learning methods do not consider the task information and only focus on the modality alignment, remaining an area to be explored.} 

In the following, we list potential directions for future research. 
\begin{itemize}
    \item \textit{Feature Representation Extractor}: Although the use of LLMs as feature representation extractors has shown promise in enhancing RL, several challenges persist in future research. \rev{Short-term goals include developing a computationally efficient feature extractor and improving the generalization of LLM-derived representations. In the long term, researchers should focus on exploiting task compositionality for better generalization and creating adaptive extraction methods for diverse control tasks.}
    \item \textit{Language Translator}: \rev{Current existing works are still limited. Short-term objectives involve exploring LLMs' ability to handle more tasks and improving translation efficiency and accuracy in RL contexts. Long-term goals include developing multimodal translation capabilities and integrating these with RL algorithms, achieving a more general translator, and helping agent learning.}
\end{itemize}

\section{LLM as Reward Designer}
\label{sec:reward-designer}
The reward signal is the most important information to instruct agent learning in RL~\cite{sutton2018reinforcement}. However, despite the fundamental importance, high-performing reward functions are known to be notoriously difficult to design~\cite{eschmann2021reward}. \revv{First}, specifying human notions of desired behavior is difficult via designed reward functions or requires huge expert demonstrations. \revv{Moreover}, dense rewards that accurately provide learning signals require either manually decomposing the general goal into sub-goals~\cite{andreas2017modular} or rewarding interesting auxiliary objectives~\cite{mirchandani2021ella}. Nevertheless, both of these methods suffer from the need for expert input and meticulous manual crafting~\cite{booth2023perils}. 

Benefiting from pre-trained common-sense knowledge, code generation, and in-context learning ability, LLM has the potential to design or shape reward functions for DRL by leveraging natural language-based instructions and environment information. In this section, we review recent literature in which LLMs act as reward models that implicitly provide reward values or explicitly write executable reward function codes detailing the calculation process of reward scalars.

\begin{figure}[bthp]
    \centering
    \includegraphics[width=\linewidth]{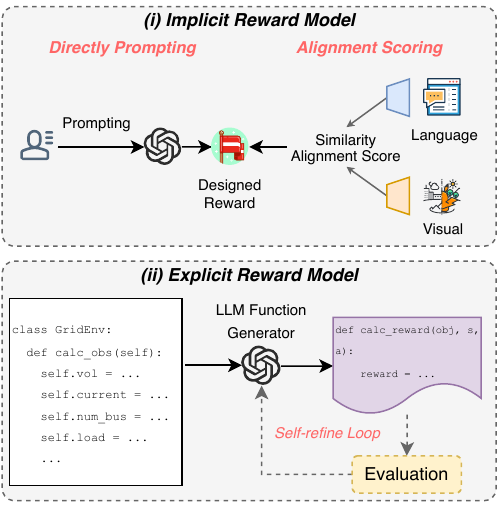}
    \caption{LLM as a reward designer. \rev{(i) Implicit Reward Model: LLMs provide rewards through direct prompting or alignment scoring between language instructions and visual observations. (ii) Explicit Reward Model: LLMs generate executable code for reward functions, with potential for self-refinement through evaluation loops.}}
    \label{fig:reward-designer}
\end{figure}

\subsection{Implicit Reward Model}
A large pre-trained model can be an implicit reward model that directly provides auxiliary or overall reward value based on the understanding of task objectives and observations. The methods are illustrated in Fig.~\ref{fig:reward-designer} (i). One way is by directly prompting with language descriptions and another way is by scoring the alignment between the feature representation of the visual observations and language-based instructions.

\subsubsection{Direct Prompting}
Reference~\cite{kwonRewardDesignLanguage2022} simplified the manual reward design by prompting an LLM as a proxy reward function with examples of desirable behaviors and the preferences description of the desired behaviors. Reference~\cite{wuReadReapRewards2023} proposed a Read and Reward framework that utilizes LLMs to read instruction manuals to boost the learning policies of specific tasks. The framework includes a \rev{Question \& Answer (QA)} extraction module for information retrieval and summarization and a Reasoning module for evaluation. Experimentally, they show RL algorithms, by their design, can obtain significant improvement in performance and training speed. In reference~\cite{cartaGroundingLargeLanguage2023}, Carta \textit{et al.} proposed an automated reward shaping method where the agent extracts auxiliary objectives from the general language goal. Using a question generation and \rev{QA system}, the framework guides the agent in reconstructing partial information about the global goal and provides an intrinsic reward signal for the agent. This intrinsic reward incentivizes the agent to produce trajectories that help them reconstruct the partial information about the general language goal. To acquire a generalizable policy by continually learning a set of tasks is challenging for RL agents since the agent is required to retain the previous knowledge while quickly adapting to new tasks. Reference~\cite{chuAcceleratingReinforcementLearning2023} introduced the Lafite-RL (Language agent feedback interactive RL) framework that provides interactive rewards mimicking human feedback based on LLMs' real-time understanding of the agent's behavior. By designing two prompts, one to let LLM understand the scenario and the other to instruct it about the evaluation criterion, their framework can accelerate the RL process while freeing up human effort during the interaction between agent and environment. 

\begin{algorithm}[hbtp]
\caption{\revv{Language Reward Modulated Pretraining (LAMP) in~\cite{LanguageRewardModulation2023}}}
\label{alg:LAMP}
\begin{algorithmic}[1]
\STATE Initialize parameters for Masked World Models (MWM)
\STATE Load pretrained DistilBERT~\cite{sanh2020distilbertdistilledversionbert} for language preprocessing
\STATE Load pretrained R3M visual encoder and score predictor
\STATE Initialize empty replay buffer
\STATE Populate language prompt buffer and synonym buffers with predefined samples
\FOR{each training episode}
    \STATE Randomize scene textures from Ego4D and RLBench datasets
    \STATE Sample ShapeNet objects and language prompts
    \STATE Insert ShapeNet objects into the scene
    \STATE Generate and process language embeddings using DistilBERT
    \STATE Execute policy to collect transitions
    \STATE Assign LAMP rewards using R3M score predictos
    \STATE Update buffers and train MWM with augmented rewards
\ENDFOR
\end{algorithmic}
\end{algorithm}

\subsubsection{Alignment Scoring}
\label{subsubsec:alignment-scoring}
For visual RL, some literature utilizes vision-language models as reward models to align multimodal data and calculate the similarity score using metrics such as cosine similarity. Rocamonde \textit{et al.} employs the CLIP model as a zero-shot reward model to specify tasks via natural language~\cite{rocamondeVisionLanguageModelsAre2023}. They first compute the probability $p_{o_t,l}$ that the agent achieves a goal given by language description $l$ out of a set of potential goals $l' \in \mathcal{L}$ in the task set $\mathcal{L}$ using the softmax computation with temperature $\tau$ over the cosine similarity between visual state embeddings $f_{\theta}(o_t)$ and language description embeddings $g_\theta(l)$ across the set of potential goals $l'$:
\begin{equation}
p_{o_t,l} = \frac{\exp(f_\theta(o_t) \cdot g_\phi(l) / \tau)}{\sum_{l'} \exp(f_\theta(o_t) \cdot g_\phi(l') / \tau)}.
\end{equation}
Then the reward is obtained by a binary reward function $r_t = r(o_{t+1}, l) = \mathbb{I}[p_{o_{t+1},l} \geq \beta]$, which thresholding the probability. Their framework only requires a single-sentence text prompt description of the desired task with minimal prompt engineering. Another work~\cite{kimGuideYourAgent2023} constructed reward signals based on the similarity between natural language-based description and embeddings from the pre-trained VLM encoder. By labeling the expert demonstration with the reward signals, the framework effectively mitigates the problem of mis-generalization. \revv{In reference~\cite{LanguageRewardModulation2023}, the authors proposed the Language Reward Modulated Pretraining (LAMP) framework as a pertaining utility for RL as opposed to a downstream task reward to warm-start sample-efficient learning. The framework leverages frozen, pre-trained VLMs such as R3M~\cite{Rev2nair2022r3muniversalvisualrepresentation} to generate noisy, albeit shaped exploration rewards by computing the alignment score between instructions and image observations.} \revv{The algorithm is presented in Algorithm~\ref{alg:LAMP}. They used Masked World Model~\cite{seoMaskedWorldModels2023}, a visual model-based RL algorithm for robot manipulation based on image and instructions. The images were downloaded from Ego4D~\cite{Rev2grauman2022ego4d} and the language instructions were obtained by querying ChatGPT. The reward is then calculated from the R3M alignment score. After that, the generated rewards are optimized with standard novelty-seeking exploration rewards for language-conditioned policy.} \rev{Reference~\cite{wang2024rlvlmfreinforcementlearningvisionRev} explored preference-based RL, where the agent learns a reward function from preference labels over the behaviors. A VLM is leveraged to generate preference labels given visual observations and a text description of the task goal. The evaluation is based on a series of vision-based manipulation tasks. Results suggested that prompting VLMs to produce preference labels for reward learning leads to better performance, in contrast to treating them as reward functions to produce raw reward scores.}

\subsection{Explicit Reward Model}
\label{subsec:explicit-reward-model}
Another way to design reward functions is by generating executable codes that explicitly specify the details of the calculation process, as illustrated in Fig.~\ref{fig:reward-designer} (ii). Compared to the implicit reward value provision, this explicit way transparently reflects the reasoning and logical process of LLMs and thus is readable for humans to further evaluate and optimize.

To help robots learn low-level actions, reference~\cite{yuLanguageRewardsRobotic} harnessed the code generation ability of LLMs to define the lower-level reward parameters based on high-level instructions. Using such a reward design paradigm, their work bridges the gap between high-level language instructions to low-level robot actions and can reliably tackle 90\% of the designed tasks compared to the 50\% of the baseline. Motivated by the capability of LLM for self-refinement~\cite{madaan2023selfrefine}, reference~\cite{songSelfRefinedLargeLanguage2023} proposed a framework with a self-refinement mechanism for automated reward function design, including initial design, evaluation and self-refinement loop. Their results indicate that the LLM-designed reward functions are able to rival or surpass manually designed reward functions. Similarly, \textit{Eureka}~\cite{maEurekaHumanLevelReward2023} developed a reward optimization algorithm with self-reflection. \revv{The algorithm is outlined in Algorithm~\ref{alg:eureka}. In each iteration, it uses an environment source code and task description to sample different reward function candidates from a coding LLM. Then the candidates are used to instruct RL training. After training, the results are used to calculate the scores of the reward candidates. Then the best reward function code are selected for reflection, where LLM uses the reasoning capability to progressively improve the reward code.} In the experiment, results show that their proposed method can achieve human-level performance on reward design and solve dexterous manipulation tasks that were previously infeasible by manual reward engineering. Another work Text2Reward~\cite{Text2RewardRewardShaping2023} generated shaped dense reward functions as executable programs based on the environment description. Given the sensitivity of RL training and the ambiguity of language, the RL policy may fail to achieve the goal. Text2Reward addresses the problem by executing the learned policy in the environment, requesting human feedback and refining the reward accordingly.

\begin{algorithm}
    \caption{\revv{Eureka~\cite{maEurekaHumanLevelReward2023}}}
    \label{alg:eureka}
    \begin{algorithmic}[1]
    \REQUIRE Task description $l$, environment code $C$, coding LLM $M_c$, evaluation function $E$, initial prompt \texttt{prompt}, optimization iterations $N$, iteration batch size $K$
    \FOR{$i \leftarrow 1$ \TO $N$}
        \STATE \texttt{// Sample $K$ reward code candidates from coding LLM $M_c$}
        \STATE $R_1, \dots, R_K \leftarrow \texttt{sample}(l, M_c, C, \texttt{prompt})$
        \STATE \texttt{// Evaluate reward candidates}
        \STATE $s_1^i = E(R_1^i)$, $s_2^i = E(R_2^i)$, \dots, $s_K^i = E(R_K^i)$
        \STATE \texttt{// Select the best reward code}
        \STATE ${\mathrm{best}} = \arg\max_k(s_1^i, \dots, s_K^i)$
        \STATE \texttt{// Reward reflection}
        \STATE \texttt{prompt}$\leftarrow$ \texttt{prompt:Reflection}($R^i_{\mathrm{best}}, s^i_{\mathrm{best}}$)
        \STATE \texttt{// Optimize Eureka reward code}
        \IF{$s^i_{\mathrm{best}} > s_{\mathrm{Eureka}}$}
            \STATE $R_{\mathrm{Eureka}}, s_{\mathrm{Eureka}} \leftarrow (R^i_{\mathrm{best}}, s^i_{\mathrm{best}})$
        \ENDIF
    \ENDFOR
    \RETURN $R_{\mathrm{Eureka}}$
    \end{algorithmic}
\end{algorithm}

\subsection{Summarization and Outlook}
\label{subsec:summarization-reward}
The use of LLMs as reward designers in RL offers a more natural and efficient way to design complex reward functions. By leveraging natural language processing capabilities, LLMs simplify the traditionally challenging task of reward function design, enhancing both the efficiency and effectiveness of RL algorithms. 

\revv{
However, the inherent biases in LLMs may transfer to the designed reward functions, potentially resulting in suboptimal or harmful behaviors~\cite{Rev2ferrara2023should}. This presents a potential risk requiring careful consideration. While existing mitigation efforts primarily address biases in demographic, cultural, and political beliefs~\cite{Rev2gallegos2024bias}, task-specific biases remain understudied, thus limiting the applicability of LLM-designed reward functions. Toward this end, we list some potential solutions. First, in preference-based RL, when an RL agent optimizes against biased reward models generated by LLMs to predict human preferences, overoptimization and overfitting may occur~\cite{Rev2pmlr-v202-gao23h}, impeding the learning of the true reward function. Reward function regularization~\cite{Rev2chakraborty2023rebel} includes the agent preference generated by the value function as a regularization term to mitigate the risk, which helps recover the true underlying reward function. Secondly, human-in-the-loop approaches are another direction to prevent harmful behaviors from designed reward functions. One viable solution is to design an evaluation module where humans can intervene and correct undesired behaviors and reward functions when the agent's action violates a predefined set of rules. Finally, principles from ensemble learning suggest that combining the reward functions from different LLM models mitigates bias from individual LLMs, leading to improved unbiased performance compared to a single LLM~\cite{Rev2radwan2024addressing}.
}

In the future, we expect advancements in the following areas:
\begin{itemize}
    \item \textit{Implicit Reward Model}: \rev{An immediate focus may consider improving how well LLM-generated rewards align with human intentions. This involves refining the quality of language instructions to reduce ambiguities and inaccuracies, ensuring that the reward functions align precisely with human notions of desired behavior. In the longer term, generalization and transferability of LLM-generated rewards across different tasks and environments, especially in complex, high-dimensional visual environments, is also an important direction to explore.}
    \item \textit{Explicit Reward Model}: \rev{A key limitation in reward code generation is its dependency on pre-trained common-sense knowledge, which can be restrictive for highly specialized tasks not covered in training data. Therefore, short-term goals may include enhancing prompts with detailed, task-specific information and external knowledge. Additionally, the reliance on manually designed templates for motion descriptions limits the adaptability. Looking further ahead, researchers might develop automated or unified processes for designing templates, moving beyond the current limitations of manual motion description templates.}
\end{itemize}

\section{LLM as Decision-Maker}
\label{sec:decision-maker}
LLMs trained on a massive amount of data show impressive results in language understanding tasks~\cite{brown2020language}, instruction-following~\cite{inoue2022prompter}, vision-language navigation~\cite{majumdar2020improving} and tasks requiring planning and sequential reasoning~\cite{ahnCanNotSay2022}. Such success motivates researchers to explore the potential of LLMs for decision-making problems. \rev{In this section, we divide the role of LLM as 1) the \textit{action-maker} that generates actions; and 2) the \textit{action-guider} that instructs the actions.} 

\begin{figure}[bhtp]
    \centering
    \includegraphics[width=\linewidth]{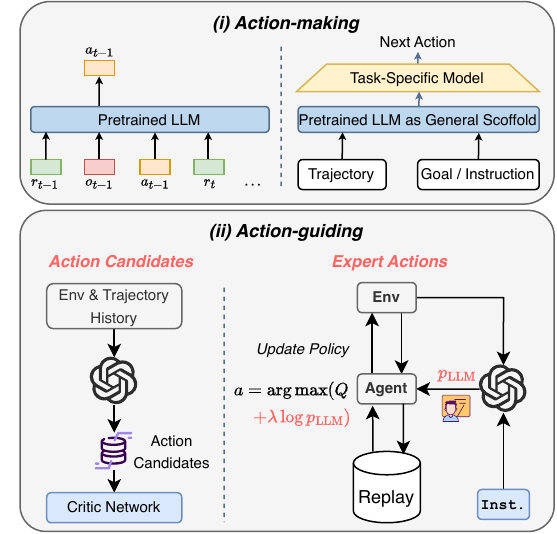}
    \caption{LLM as a decision-maker. \rev{(i) Action-Making: given a $T$-length trajectory $\tau = (\hat{R}_{1}, s_{1}, a_{1}, \dots, \hat{R}_{T}, s_{T}, a_{T})$ as a sequence of ordered return-to-go $\hat{R}$, action $a$, and states $s$, LLM learns to predict future action $a'_t$ by minimizing the mean squared error loss $\mathcal{L} = \sum_{t}\left\|a_t-a_t^{\prime}\right\|_2^2$. (ii) Action-Guiding: LLM generates a reduced set of action candidates for agents or generates expert actions to regularize RL learning.}}
    \label{fig:decision-maker}
\end{figure}

\subsection{\rev{Action-Making}}
\label{subsec:action-making}
\rev{Transformer-based models such as Decision Transformer (DT)~\cite{janner2021offline} have shown great potential in offline RL domain. Instead of using the traditional trial-and-error way, these models treat offline RL as a sequence modeling problem, yielding promising results. As LLM itself is a large-scale Transformer-based model, a natural thought is to leverage the pre-trained power of LLM within this paradigm.}

\rev{We term this function of LLM as action-making, and identify two typical approaches, as illustrated in Fig.~\ref{fig:decision-maker} (i). In the first approach (left figure), pre-trained LLM is fine-tuned and then employed for action generation. In the second approach (right figure), goal/instruction along with trajectory are fed to pre-trained LLM. Moreover, a task-specific smaller model is appended after the fine-tuned LLM to facilitate rapid adaptation to diverse tasks.}

\rev{Pre-trained LLMs outperform basic DT in generalization and sample efficiency, especially for sparse-reward and long-horizon tasks. LLMs' latent representations, learned from diverse linguistic data, provide valuable prior knowledge for new tasks. This knowledge enables LLMs to solve unseen tasks with less training data, by transferring knowledge from similar tasks and predicting high-reward actions even with sparse feedback. For instance, comparing pre-trained LLM with basic DT, Li et al.~\cite{liPreTrainedLanguageModels2022} reported a 43.6\% improvement in out-of-distribution (novel) task completion rates while requiring less training data (e.g., 500 vs 10K). Additionally, Shi et al.~\cite{shiUnleashingPowerPretrained2023} demonstrated a 50\% performance gain in sparse-reward environments like Kitchen and Reacher2d. For long-horizon tasks, pre-trained representations encode future information, guiding decision-making over extended sequences. For instance, in AntMaze, a long-horizon navigation environment, pre-trained representations yield five times higher scores compared to non-pre-trained counterparts~\cite{zeng2024goal}.}

\rev{Several studies have further explored the application of LLMs in offline RL, demonstrating their versatility and effectiveness across various tasks and benchmarks.} \rev{Reference~\cite{reidCanWikipediaHelp2022} investigated the transferability of general language models on specific RL tasks. Fined-tuned on offline RL tasks (control, games), these general language models outperform Decision Transformer and reduce training time by 3-6x on D4RL benchmark~\cite{fu2020d4rl}}. \rev{Reference~\cite{liPreTrainedLanguageModels2022} used pre-trained LLM as a general scaffold for task-specific model learning, where goals were added along with observations as the input for LLM. Results on embodied decision-making tasks demonstrate that their proposed method outperforms others with less training data, especially when generalizing to novel tasks. In addition, they found representations in pre-trained language models can aid learning and generalization even outside of language.} To unify language reasoning with actions in a single policy, reference~\cite{mezghaniThinkYouAct2023} generated textual captions interleaved with actions when training the Transformer-based policy. Results show that by using captions describing the next subgoals, the reasoning policy can consistently outperform the caption-free baseline. \revv{\label{rev:unleash}For scenarios where data collection is costly and risky, reference~\cite{shiUnleashingPowerPretrained2023} proposed a general framework to effectively use pre-trained LLM for offline RL. The pre-trained LLM are based on DT. To combine the pre-trained knowledge and task-related domain knowledge, they fine-tuned pre-trained LLM with Low-Rank Adaptation (LoRA) method. The architecture of Transformer is based on GPT-2 model with 12 layers and 12 attention heads. To fine-tune the LLM, they obtained the trajectory data from the D4RL dataset.} Results show their method achieves state-of-the-art performance in sparse-reward tasks with limited data samples. To integrate multimodal data, e.g., vision and language, into the offline RL, reference~\cite{brohanRT2VisionLanguageActionModels} co-fine-tuned vision-language models on both robotic trajectory data and Internet-scale vision-language tasks, e.g., visual question answering. In the framework, they incorporate the actions as natural language tokens and co-fine-tune models on both vision and language datasets. Results show that such co-fine-tune methods can increase generalization performance and the chain of thought reasoning can help the agent perform multi-stage semantic reasoning and solve complex tasks.

\subsection{\rev{Action-Guiding}}
The \rev{action-guider} role is illustrated in Fig.~\ref{fig:decision-maker} (ii). As an \rev{action-guider, LLM guides the action selection by either generating reasonable action candidates or expert actions. By instructing the action selection, LLM improves sample efficiency and exploration efficiency posed by enormous action spaces and natural language.}

\subsubsection{Action Candidates}
\rev{In environments such as text-based games, action spaces are large and only a tiny fraction of actions are accessible. Although RL agents can learn through extensive trials, they often face exploration efficiency issues, especially in multi-task settings where agents must manage various tasks simultaneously. LLMs address this challenge by generating a reduced set of action candidates based on task understanding. These candidates are likely to yield high rewards and are applicable across multiple tasks, enhancing exploration efficiency and reducing the need for ineffective exploration.} Reference~\cite{yaoKeepCALMExplore2020} trained a GPT-2 model to generate the candidates. To maximize long-term rewards, another neural network is used to calculate the Q-values of these candidates. \rev{When tested in 28 man-made text games from Jericho framework~\cite{hausknecht2020interactive}, they found the method excludes non-useful actions, speeds up the exploration and consistently achieves higher scores by more than 20\%.} Following this work, another study~\cite{ahnCanNotSay2022} proposed the SayCan framework, where instruction-following robots are integrated with an embodied LLM to understand tasks. When receiving instructions, the embodied LLM generates a high-level step-by-step plan. When acting, LLM produces action candidates based on task prompts and then the candidate with the largest critic value is executed. \rev{Being evaluated in an office kitchen with real-world robotic tasks from 101 instructions, their method can complete long-horizon, abstract, natural language instructions on a mobile manipulator.}

\subsubsection{\rev{Expert Actions}}
\rev{Traditional RL agents cannot converge to desirable equilibrium in human-AI collaboration or learn efficiently for complex tasks, due to the lack of expert demonstrations. With the understanding of human behavior and general knowledge, LLM solves the issues by producing high-quality expert actions to regularize RL agents.} \rev{In human-AI collaboration}, \textit{instructRL}~\cite{huLanguageInstructedReinforcement2023} used LLM to generate prior policy based on human instructions and uses this prior to regularizing the RL objective. Specifically, \textit{instructRL} augments the policy update function with an auxiliary term $p_{\mathrm{LLM}}[\mathrm{lang(}a_t\mathrm{)|lang(}\tau_t^i\mathrm{), inst}]$, the probability of choosing an action based on the trajectory and instructions. Experiments show \textit{instructRL} converges to policies aligned with human preferences. \rev{Similarly, to address the sample inefficiency issue of RL, Zhou \textit{et al.}~\cite{zhouLargeLanguageModel2023} included the policy difference between the student model and LLM-based teacher into RL learning loss. Their experiments on simulation platforms demonstrate that the method reduced training iterations by a factor of 1 to 9.} In reference~\cite{dalal2024planseqlearnRev}, LLM was leveraged to generate a high-level expert motion plan of robotics tasks, guiding RL policies to efficiently solve robotics control tasks. The high-level language plan breaks long-horizon tasks into stages to execute. Then, a single RL policy was trained across all states and stepped through the language plan. \rev{Their results show the proposed method solves long-horizon tasks from raw visual input spanning different benchmarks at success rates of over 85\%, out-performing classical, language-based, and end-to-end approaches.}

\subsection{Summarization and Outlook}
\label{subsec:summarization-decision-maker}
\rev{Sample inefficiency and exploration inefficiency remain long-standing challenges for deep RL, particularly in environments with sparse rewards or where data collection is expensive or risky. LLM provides three ways to solve the problems. First, LLM as action-makers treat RL as a conditional sequence modeling problem. The supervised way fine-tunes pre-trained LLMs to predict future actions. Benefiting from learned prior knowledge, LLM can perform well even in out-of-distribution, sparse-reward, and long-horizon tasks. Second, LLM as action-guiders generates potential action candidates for RL. With task comprehension, these action candidates promote agents to explore potentially high task-value states, thus increasing exploration efficiency. Last, LLM generates expert actions to help RL learn from demonstrations. By incorporating demonstrations from LLM, and RL learns specific prior knowledge and improves sample efficiency.}

\revv{
Safety issues are another important topic when using LLMs as decision-makers in RL, particularly for costly and risky tasks~\cite{Rev2garcia2015comprehensive}. For action-making, DT-based offline RL learns the optimal policy from pre-collected datasets. Recently, integrating safety constraints has been explored by some works using methods such as pessimistic estimations~\cite{Rev2xu2022constraints} and stationary distribution correction~\cite{Rev2lee2022coptidice}. These methods set a constant constraint threshold before training. To dynamically adjust the threshold during deployment, a constrained decision transformer that dynamically relabels the reward based on safety-reward trade-offs was proposed~\cite{Rev2liu2023constrained}. For action-guiding, LLMs are viewed as instructors to provide expert actions. Ensuring the safety of actions generated by LLMs differs from that in action-making. Leveraging ideas from LLM-based agent research, we propose several potential solutions. First, LLMs can be equipped with testing modules that execute code to evaluate action safety and feasibility~\cite{Rev2naihin2023testing}. Second, implementing a carefully designed human-in-the-loop framework enables safety intervention through human oversight~\cite{Rev2huang2024safety}. Finally, developing memory modules with reasoning mechanisms that adaptively learn safety boundaries from past experiences provides a continual learning-based approach to ensuring long-term safety~\cite{shinnReflexionLanguageAgents2023}.
}

\rev{In the following, we list the challenges and future directions of the two roles as below:}
\begin{itemize}
    \item \textit{\rev{Action-making}}: \rev{Directly employing pre-trained large-scale LLM to generate actions demands huge computational resources and huge data for fine-tuning it in specific tasks. In the short term, future work may consider more cost-effective methods such as the Low-Rank Adaptation (LoRA)~\cite{hu2021lora} to exploit the power of LLM in direct decision-making. Long-term goals involve developing innovative techniques to efficiently exploit the power of LLMs in direct decision-making, potentially creating hybrid models that combine the strengths of LLMs with more lightweight, task-specific architectures.}
    \item \textit{\rev{Action-guiding}}: Since the LLM acts as an instructor to provide \rev{expert actions}, the bias and limitations are also inherited by the agent. In addition, LLM itself cannot inherently interrogate or intervene in the environment, which limits LLMs' capabilities to intentionally aggregate information. \rev{Short-term goal is to address the inherited biases and limitations when LLMs act as instructors providing expert actions, focusing on methods to filter or correct biased information. In the long term, it is crucial to develop mechanisms for LLMs to actively interrogate and intervene in the environment, enabling them to intentionally aggregate information and improve their capabilities through real-world interactions. Therefore, how to use the information gained from real-world interactions to improve the LLM itself in terms of actuality and reasoning is another important problem.}
\end{itemize}

\section{LLM as Generator}

The generative capability of LLMs can be applied to environmental simulation and behavior explanation. On the one hand, growing interests in applying RL to the real world and risky data-collection process suggests a need for imaginary rollouts~\cite{hafnerDreamControlLearning2020},~\cite{hafnerMasteringAtariDiscrete2022}. Possessed with a powerful modeling capability and world knowledge, LLMs can serve as a \textit{world model simulator} to learn complex environmental dynamics with high fidelity by iteratively predicting the next state and reward, thus increasing the sample efficiency in model-based RL~\cite{hafner2020dream},~\cite{matsuo2022deep}. On the other hand, in RL, interpretability remains an important security issue in current black-box AI systems as they are increasingly being deployed to help end-users with everyday tasks. Explanations of policy can improve the end-user understanding of the agent's decision-making and inform the reward function design for agent learning. In such aspects, LLM can act as a policy interpreter based on their knowledge and reasoning ability. In this section, we classify the above two roles of LLM as a generator and review recent related works.
\begin{figure}[thbp]
    \centering
    \includegraphics[width=\linewidth]{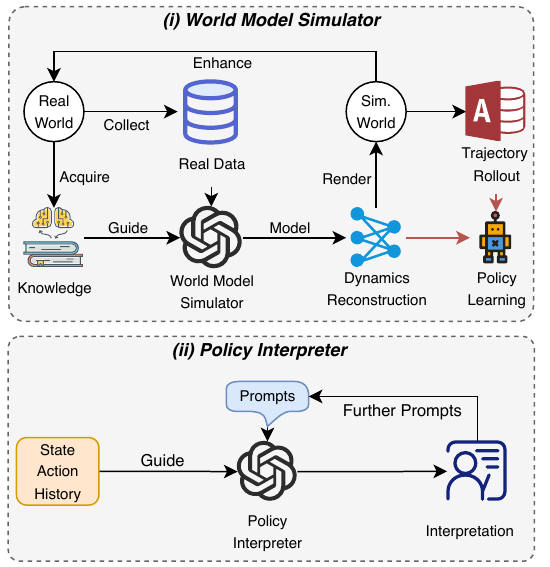}
    \caption{LLM as a generator. \rev{(i) World Model Simulator: LLM uses real-world data and knowledge to model dynamics, generate simulated worlds, and assist policy learning. (ii) Policy Interpreter: LLM generates interpretations of agent behavior based on state-action history and prompts, potentially leading to explainable RL.}}
    \label{fig:generator}
\end{figure}
\label{sec:generator}
\subsection{World Model Simulator}
Serving as a world model simulator, LLM is trained as a 1) \textit{trajectory rolloutor}, which auto-regressively generates accurate trajectories for the agent to learn and plan; 2) \textit{dynamics representation learner}, which predicts the latent representation of the world using representation learning. A flow chart of the world model is illustrated in Fig.~\ref{fig:generator} (i). From the real world, knowledge and real data can be collected to construct a world model simulator, which further models the dynamics representation of the world, generates trajectories and helps the policy learning of the agent in the real world.

\subsubsection{Trajectory Rolloutor} Similar to the decision transformer~\cite{chen2021decision} in offline RL, pre-trained large-scale models were used in model-based to synthesize trajectories. In 2022, Micheli \textit{et al.} proposed IRIS, an agent that employs a discrete autoencoder and an autoregressive Transformer to learn the world model for Atari games~\cite{micheliTransformersAreSampleEfficient2022}. With the equivalent of two hours of gameplay in the Atari 100k benchmark, the proposed method outperforms humans in 10 out of 26 games. Similarly, reference~\cite{robineTransformerbasedWorldModels2023} applied a transformer to build a sample-efficient world model for Atari games. Utilizing such a Transformer-based world model (TWM), the RL agent can solve the long-term dependency and train a state-of-the-art policy on the Atari 100k benchmark based on generated meaningful experiences from TWM. Visual RL enables the RL agent to learn from visual observations effectively. Reference~\cite{chenTransDreamerReinforcementLearning2022} proposed TransDreamer, a Transformer-based model-based RL agent that leverages a Transformer for dynamics predictions in 2D and 3D visual RL tasks. Experiments showed the TransDreamer agent can outperform Dreamer with long-range memory access and memory-based reasoning. Based on the development of supervised pre-training methods, Seo \textit{et al.} proposed a framework that learns world model dynamics from action-free video representations~\cite{seoReinforcementLearningActionFree2022}. The framework added a video-based intrinsic bonus for better-guiding exploration to effectively encourage agents to learn diverse behaviors. The experimental results demonstrated their proposed method can improve the performances of vision-based RL on manipulation and locomotion tasks by transferring the pre-trained representations from unseen domains. 

\subsubsection{Dynamics Representation Learner}
\label{subsubsec:dynamic-representation-learner}
Using representation learning techniques, the latent representation of the future can be learned to assist decision-making. Reference~\cite{seoMaskedWorldModels2023} introduced a visual model-based RL framework that decouples visual representation learning and dynamic learning by training an autoencoder with a vision Transformer to reconstruct pixel-given masked observations and learn the dynamics from the latent space. By such a decoupling approach, their proposed method achieved state-of-the-art performance on visual robotic tasks from Meta-world and RLBench. \revv{Utilizing the fact that language contains rich information signals and can help agents to predict the future, Lin \textit{et al.} proposed Dynalang~\cite{linLearningModelWorld2023}, where an agent that learns a multimodal world model to predict future text and image representations and thereby instruct the decision-making process. The algorithm is presented in Algorithm~\ref{alg:dynalang}. In the training part, a LLM-based world model implemented with Recurrent State Space Model (RSSM)~\cite{Rev2hafner2019learning} computes the state representation $z_t$ based on the collected transitions. After that, the representation $z_t$ and the action $a_t$ are fed into the RSSM to predict the future representations $z_{t+1}$. In addition, the language $\hat{l}_t$ and images $\hat{x}_t$ are predicted to reconstruct the original state. Then, the world model is trained to learn the next representation and reconstruct observations from the representations. After updating, the world model imagined (sampled) rollouts and the policy is trained to maximize the imagined rewards. Compared to other works that use language only for predicting actions, multimodal information enables Dynalang to handle tasks that require grounded language generation, obtaining higher score than methods provided with only task descriptions.} To solve the out-of-distribution generalization problem in visual control tasks of RL, reference~\cite{poudelLanGWMLanguageGrounded2023} proposed the Language Grounded World Model (LanGWM), which focuses on learning language-grounded visual features to enhance the world model learning. To improve the generalization of the learned visual features, they masked the bounding boxes and predicted them with given language descriptions. Utilizing the expressing ability of language in higher-level concepts and global contexts, the proposed LanGWM method yields state-of-the-art results on out-of-distribution tests. 

\begin{algorithm}[hbtp]
    \caption{\revv{Training part of Dynalang~\cite{linLearningModelWorld2023}}}
    \label{alg:dynalang}
    \begin{algorithmic}[1]
    \REQUIRE Rewards $r_t$, episode continue flag $c_t$, images $x_t$,
        language tokens $l_t$, actions $a_t$, model state $(h_t, z_t)$.
    \WHILE{training}
        \STATE Draw batch of transitions $\{(r_t, c_t, x_t, l_t, a_t)\}$ from replay buffer.
        \STATE Use world model to compute multimodal representations $z_t$, future predictions $\hat{z}_{t+1}$, and decode $\hat{x}_t$, $\hat{l}_t$, $\hat{r}_t$, $\hat{c}_t$.
        \STATE Update world model to minimize $\mathcal{L}_{\text{pred}} + \mathcal{L}_{\text{repr}}$.
        \STATE Imagine rollouts from all $z_t$ using $\pi$.
        \STATE Update actor to minimize $\mathcal{L}_{\pi}$.
        \STATE Update critic to minimize $\mathcal{L}_{V}$.
    \ENDWHILE
    \end{algorithmic}
    \end{algorithm}

\subsection{Policy Interpreter}
Explainable RL (XRL) is an emerging subfield of both explainable machine learning and RL that has attracted considerable attention recently. XRL aims to elucidate the decision-making process of learning agents. According to a survey in explainable RL~\cite{milani2022survey}, the categories of XRL include the feature importance, learning process and MDP, and policy level. Currently, the usage of LLMs in XRL has only been limited to the policy level, \textit{i.e.}, as the policy interpreter. Therefore, though there is limited literature related to this area, we think this field is important and requires more attention in the future. The following will introduce the role of LLMs as policy interpreters and an outlook regarding other categories of XRL will be provided in the last subsection.

As the policy interpreter, LLM generates explanations with the prompts of state and action description or trajectory information. An illustration is depicted in Fig.~\ref{fig:generator}. Using the trajectory history of states and actions as context information, LLMs can be prompts to generate readable interpretations of current policies or situations for humans.

Das \textit{et al.}~\cite{das2023stateexplanation} proposed a unified framework called State2Explanation (S2E), that learns a joint embedding model between state-action pairs and concept-based explanation. Based on the learned models, the explanation can help inform reward shaping during an agent's training and provide insights to end-users at deployment. Another work ~\cite{UnderstandingLanguageWorld2023} first distilled the policy into a decision tree, derives the decision path, and then prompts an LLM to generate a natural language explanation based on the decision path. \rev{Additionally, \textit{Lu et al.}~\cite{lu2023closerRev} introduced a framework that decomposes the overall reward into multiple sub-rewards based on specific object properties, defines actions as high-level motion primitives executed at precise 3D positions to simplify decision-making, and integrates LLMs to enable interactive and flexible querying of explanations.}

\subsection{Summarization and Outlook}

As a generator, LLMs can be integrated into model-based RL or explainable RL, \textit{i.e.}, serving as world model simulators or policy interpreters, respectively. As world model simulators, LLMs enhance model-based RL by auto-regressively generating accurate trajectories (trajectory rollout) and by predicting latent world representations (world representation learners), significantly improving sample efficiency and decision-making accuracy. In the realm of explainable RL, LLMs provide valuable insights for both end-user understanding and reward shaping by generating explanations based on trajectory information. However, the current usages of LLMs in explainable RL are rather limited and have great potential for future work. Below are discussions on the key limitations and future work directions of the two roles:
\begin{itemize}
    \item \textit{World Model Simulator}: LLMs encounter challenges in aligning their abstract knowledge with the specific requirements of different environments, leading to limitations in functional competence and grounding. This misalignment affects their effectiveness in generating trajectories and interacting with the environment. Additionally, the current model-based agents, which rely on purely observational world models, present difficulties for human adaptation. These models are typically modifiable only through observational data, which is not an effective means for humans to communicate complex intentions or adjustments. \rev{Furthermore, while LLMs hold promise for enhancing multi-task learning by generating trajectories and dynamic representations applicable to multiple tasks, there remains a scarcity of research exploring this potential.Looking ahead, in the short term, researchers might prioritize improving LLMs' alignment with specific environment requirements. For the long term, a promising direction would be to integrate language instructions or an adapter into the world model. This approach could lead to a more flexible and adaptive world model that can better accommodate human intentions and adjustments, as well as support more effective multi-task learning through enhanced trajectory generation and dynamic representations.}
    \item \textit{Policy Interpreter}: As policy interpreters, the quality of explanations depends on the LLM's understanding of feature representations and implicit logic of policy. How to utilize domain knowledge or examples to improve the understanding of a complex policy is still a major issue. \rev{In the short term, researchers could focus on enhancing LLMs' ability to interpret the correlation between observations and policy selection, providing insights into the decision-making process of RL agents. This could involve developing techniques to better leverage domain knowledge and examples for improving LLMs' understanding of complex policies. In the long-term, the field could explore advanced applications of LLMs in explainable RL, such as analyzing the learning process and MDP to reveal influences on agent behavior. Researchers could develop sophisticated prompting techniques for LLMs to answer nuanced ``why'' and ``why-not'' questions with MDP context, providing deeper explanations of agent decision-making.}
\end{itemize}

\section{Discussion}
\label{sec:discussion}
\rev{In previous sections, we introduced the concept ``LLM-enhanced RL'' and developed a corresponding framework, which we extended to the integration of multimodal AI models such as visual-language models. We then discussed the different LLM-enhanced RL approaches.}
\rev{This section provides a comprehensive analysis of the LLM-enhanced RL approach. We begin with a comparative analysis of the different LLM roles, highlighting their advantages and limitations. Following this, we explore potential real-world applications of the LLM-enhanced RL paradigm. Finally, we discuss future opportunities and challenges, taking into account both the untapped capabilities and inherent limitations of LLMs, with a particular focus on their application in multimodal information environments.}
\subsection{\rev{Comparison of Different LLM-Enhanced RL Approaches}}
\rev{This section provides a comparative analysis of the four LLM-enhanced RL approaches, helping researchers understand the strengths and limitations of each approach.}
\begin{itemize}
\item \rev{\textbf{Information Processor}: As an information processor, LLM excels in handling complex, multimodal inputs, particularly in translating natural language instructions or environment information into a format more readily usable by RL agents. This role significantly enhances the agent's ability to understand and interact with complex environments. However, it faces challenges in computational efficiency and may struggle with highly specialized domain knowledge not covered in its pre-training.}
\item \rev{\textbf{Reward Designer}: LLMs serving as reward designers offer a more intuitive and flexible approach to defining reward functions, especially in complex or sparse-reward environments. This role can significantly improve the alignment of RL objectives with human intentions. The main limitation lies in ensuring that generated rewards accurately reflect task-specific nuances and long-term goals, particularly in highly specialized domains.}
\item \rev{\textbf{Decision-Maker}: As decision-makers, LLMs can either directly generate actions or guide action selection, leveraging their vast knowledge base to improve sample efficiency and exploration in RL. This role is particularly effective in tasks requiring complex reasoning or long-term planning. However, it may face challenges in real-time decision-making scenarios due to computational overhead and may inherit biases present in the LLM's training data.}
\item \rev{\textbf{Generator}: In the generator role, LLMs can simulate complex environments for model-based RL and provide interpretable explanations of RL policies. This capability is invaluable for improving sample efficiency and making RL more transparent and understandable. The main challenges include aligning generated simulations with real-world dynamics and ensuring the relevance and accuracy of policy explanations.}
\end{itemize}
\subsection{Applications of LLM-Enhanced RL}
Based on the characteristics of LLM-enhanced RL, such as multimodal information understanding and multi-task learning and generation, we believe that LLM-enhanced RL opens up a wide array of potential applications. Here we list several applications to inspire researchers.
\begin{itemize}
    \item \textit{Robotics}: RL is widely used in robots to learn how to make decisions and execute actions to achieve goals. Utilizing natural language understanding and general logical reasoning abilities, LLM-enhanced RL can 1) improve the efficiency of human-robot interaction, 2) help robots understand human needs and behavioral logic, and 3) enhance decision-making and planning capabilities.
    \item \textit{Autonomous Driving}: Autonomous driving uses RL to make decisions in complex, ever-changing environments that involve understanding both sensor data (visual, lidar, radar) and contextual information (traffic laws, human behavior). LLM-enhanced RL could employ LLMs to 1) process this multimodal information and natural language instructions; or 2) design comprehensive rewards based on multi-disciplinary metrics such as safety, efficiency, and passenger comfort. 
     \item \textit{Energy Management}: In the energy system, operators or users apply RL to efficiently manage the usage, transportation, conversion and storage of multiple energy with high uncertainty brought by renewable resources. LLM-enhanced RL in such cases can 1) improve the RL agents' ability to handle multi-objective tasks, such as economy, safety, and low carbon, by reward function designing, and 2) increase the sample efficiency for the new energy system.
    \item \textit{Healthcare Recommendation}: RL is used to learn recommendations or suggestions in healthcare~\cite{yu2021reinforcement}. LLMs can utilize the domain knowledge to analyze the vast amount of patient data and medical histories, therefore, 1) accelerating the learning process of RL recommendation and 2) providing more accurate diagnostic and treatment recommendations in healthcare.
   \end{itemize}

\subsection{Opportunities for LLM-Enhanced RL}
\label{subsec:opportunities}

Although current works in LLM-enhanced RL have already shown better performance in several aspects, more unexplored areas remain to be explored and may lead to significant improvement. Below, we summarize the potential opportunities of LLM-enhanced RL from both the perspectives of RL and LLM capabilities, respectively.
\begin{itemize}
    \item \textit{RL}: Existing work such as~\cite{huLanguageInstructedReinforcement2023,mezghaniThinkYouAct2023,liPreTrainedLanguageModels2022} mainly focus on the general RL while various specialized branches of RL are still under-exploited, e.g. multi-agent RL, safe RL, transfer RL, explainable RL, multi-task RL, \revv{in-context RL and human-centric RL}. \rev{In the multi-agent area, compared to various works about multi-LLM-agent collaboration~\cite{agashe2023evaluatingRev},~\cite{kannan2023smartRev}, LLM-based multi-agent RL remains largely unexplored~\cite{slumbers2023leveragingRev}. A recent survey discussed some open research problems within this field~\cite{sun2024llmRev}.} LLM-enhanced strategies could be employed to facilitate communication and collaboration among RL agents. The natural language understanding capabilities of LLMs can be used to interpret and generate instructions or strategies among agents, enhancing cooperative behaviors or competition strategies; Safe RL could benefit from the reasoning and predictive capabilities of LLMs to design cost functions that encourage safety criteria compliance, reducing the risks of dangerous exploration; In transfer RL, LLMs can assist in identifying similarities between tasks or environments, leveraging their vast knowledge base to facilitate knowledge transfer and thus improve learning efficiency and adaptability across different tasks; \rev{For multi-task RL, LLMs enhance agents by processing diverse entity representations and aligning instructions (Information Processor), generating reduced action sets and expert actions for various tasks (Decision-Maker), and creating task-specific trajectories and dynamic representations (Generator). \revv{Recently, in-context RL has emerged as a promising field by leveraging the in-context learning capability of LLMs to solve decision-making problems~\cite{Rev2lee2024supervised}. Given a query state and an in-context offline dataset, a trained LLM exhibits both online exploration and offline conservation while avoiding extensive trial and error and is sample-efficient. Additionally, human-centric RL is another prominent field in healthcare and robotics, where AI and humans learn and communicate with each other for collaboration~\cite{Rev2buccinca2024towards}. In this setting, an LLM naturally serves as a perfect mediator to facilitate bidirectional communication through natural language interactions.}}
    \item \textit{LLM}: \revv{While LLMs have been integrated with RL in various methods, several promising directions remain to be explored to further enhance LLM capabilities for RL. It can be divided into the language model view and the language agent view as follows:
   \begin{itemize}
    \item \textbf{From the model perspective}: LLM could be enhanced by an external knowledge base and continual learning. Retrieval-augmented generation (RAG) techniques help LLM to retrieve the most relevant data in an external database, which could be used to attach knowledge in the task domains~\cite{lewis2021retrievalaugmentedgenerationknowledgeintensivenlpRev}. Continual learning techniques are also a hot field that enables models to continuously acquire new knowledge while retaining previously learned capabilities~\cite{Rev2wang2024comprehensive}. This technique allows both LLMs and RL agents to continuously evolve and adapt to new tasks or environments while maintaining their fundamental pre-trained abilities, leading to more flexible and robust learning systems.
    \item \textbf{From the agent perspective}: Equipping LLM-based agents with specialized modules---namely, planning modules, memory modules, and action modules---can significantly enhance the capabilities of LLMs~\cite{Rev2wang2024survey,Rev2cheng2024exploring}. In the planning module, multi-step reasoning techniques such as CoT-based reasoning~\cite{wei2023chainofthought} and Monte Carlo tree search (MCTS)~\cite{Rev2browne2012survey} can be used to enhance LLM's long-term planning ability, improving the long-term task decomposition ability of RL agents; for the memory module, long- and short-term human-memories and corresponding memory retrieval mechanisms provide a way to store previous experiences and learn adaptively along with the RL agents~\cite{Rev2zhang2024surveymemorymechanismlarge}; for the action module, tool integration presents another promising direction to unlock new possibilities for LLMs. External tools such as mathematical solvers~\cite{Rev2patil2023gorilla} and internet browsers~\cite{Rev2nakano2021webgpt} can augment LLMs' capabilities in RL tasks requiring complex mathematical computations and real-time information processing, respectively. Furthermore, collaboration of multiple LLM agents~\cite{Rev2talebirad2023multi} is another promising direction. With each LLM playing different roles and acquiring different information, multiple distinctive LLM agents solve complex problems by communicating and exchanging information. In the context of LLM-enhanced RL, different LLMs could potential serves different roles and work together to guiding RL. For example, in a group of three LLM agents, one LLM focuses on guiding short-term or immediate problem-solving for RL; another LLM is responsible for long-term planning for RL; the last LLM serves as a coordinator responsible for overseeing the group and adjusting horizon consideration accordingly.
   \end{itemize}}
\end{itemize}

\subsection{Challenges of LLM-Enhanced RL}
\label{subsec:challenges}

While LLM improves different issues in the RL paradigm, the success of LLM-enhanced RL is inherently tied to the capabilities and limitations of the underlying LLM. The primary concerns revolve around the \revv{inherent limitations of LLMs, adaptability of LLMs in RL environments, computational demands, and the broader implications of deploying such systems in real-world scenarios.}

\begin{itemize}
    \item \revv{\textit{Inherent Limitations of LLMs}: The effectiveness of LLM-enhanced RL systems is fundamentally constrained by the inherent capabilities and limitations of the underlying language models. The presence of systematic biases and potential hallucinations in pre-trained LLMs can significantly impact the reliability of multimodal input interpretation, potentially compromising the overall performance of the RL agent. This necessitates the development of robust evaluation frameworks to systematically characterize and delineate the capability boundaries of LLMs given specified RL contexts. Furthermore, incorporating uncertainty quantification methods aids in identifying unreliable answers, increasing the trustworthiness of LLMs' responses~\cite{lin2024generatingconfidenceuncertaintyquantificationRev}.}
    \item \revv{\textit{Adaptability of LLMs in RL Environments}: Despite the vast knowledge base of LLMs, they may struggle to adapt to specific or novel RL task environments that are not well-represented in their training data. This calls for methods to expand the task-related knowledge for LLM and ground LLMs' inherent knowledge in specific domains. In terms of expanding the task-related knowledge, RAG-related techniques attach LLMs with the external domain-knowledge without further training~\cite{lewis2021retrievalaugmentedgenerationknowledgeintensivenlpRev}. When fine-tuning LLMs, employing data augmentation techniques such as generating synthetic data is another way to expand the diversity of training scenarios and improve generalization to novel environments~\cite{tan2023inferringRev}. Additionally, continual learning mechanisms help preserve previously acquired knowledge while adapting to new information~\cite{gao2023unifiedRev}. For domain-specific knowledge grounding, approaches include training value functions to evaluate the long-term utility of LLM instructions~\cite{ahnCanNotSay2022}. Expert trajectories provide another valuable source of information, allowing LLMs to learn optimal decision-making patterns through in-context learning.}
    \item \textit{\revv{Computational Demands}}: \rev{The integration of LLMs into the RL learning process introduces complexities in terms of computational overhead and inference times, which slow down the learning process of RL. To address this challenge, several potential solutions have been proposed across different levels. At the data level, input compression techniques such as prompt pruning~\cite{zhou2023efficientpromptingdynamicincontextRev} can be employed to directly shorten the model input, significantly reducing inference time without substantial loss in performance. At the model level, efficient architecture designs like mixture-of-experts (MoE)~\cite{gao2022parameterefficientmixtureofexpertsarchitecturepretrainedRev} enable conditional computation, activating only relevant parts of the model for each input, thus reducing computational costs. Similarly, structured state space models (SSM)~\cite{gu2024mambalineartimesequencemodelingRev} offer linear-time complexity for sequence modeling, providing a more efficient alternative to traditional attention mechanisms. At the system level, advanced caching strategies~\cite{delcorro2023skipdecodeautoregressiveskipdecodingRev} and asynchronous processing techniques~\cite{chen2024asynchronouslargelanguagemodelRev} can be implemented to reuse intermediate computations and parallelize the execution, respectively.}
    \item \textit{Ethical, Legal, and Safety Concerns}: In practical usages, the use of LLMs involves complex ethical, legal, and safety concerns. Data privacy, intellectual property, and accountability for AI decisions should be carefully discussed. \rev{To address these issues, researchers are developing robust frameworks for responsible AI deployment. At the privacy level, differential privacy techniques~\cite{charles2024finetuninglargelanguagemodelsRev} are being implemented to protect individual data during training and inference. For transparency, efforts focus on developing interpretable AI systems, allowing stakeholders to audit AI-driven decisions~\cite{cambria2024xaimeetsllmssurveyRev}. To enhance system robustness, adversarial training methods are being explored to strengthen LLM-RL systems against potential attacks~\cite{xhonneux2024efficientadversarialtrainingllmsRev}. Regarding ethical and legal issues, a recent survey also analyzed potential solutions to integrate ethical standards and societal values into LLM ~\cite{deng2024deconstructingethicslargelanguageRev}.}
\end{itemize}

\section{Conclusion}
\label{sec:conclusion}
LLMs, with their pre-trained knowledge bases and powerful capabilities such as reasoning and in-context learning, present as a viable solution to enhance RL in terms of natural language understanding, multi-task generalization, task planning, and sample efficiency. In this survey, we have defined this paradigm as \textit{LLM-enhanced RL} and summarized its characteristics, together with opportunities and challenges. To formalize the research scope and methodology of LLM-enhanced RL, we propose a structured framework to systematically categorize the roles of LLM based on the functionalities within the classical agent-environment interaction paradigm. According to functionalities, we categorize the roles of LLMs into information processor, reward designer, decision-maker, and generator. For each category, we review current literature based on their methods and applications, outline the methodologies, discuss the addressed issues, and provide insights into future directions. These are listed below:
\begin{itemize}
    \item \textit{Information Processor}: LLMs extract observational representations and formal specification languages for RL agents, thus increasing the sample efficiency. Future directions include incorporating goal-based information and integrating multimodal environmental information to obtain stronger information extraction ability.
    \item \textit{Reward Designer}: For intricate or un-quantifiable tasks, LLMs can leverage pre-trained knowledge to generate high-performing rewards that are notoriously difficult for humans to design. Current methods still require consistently modifying the prompts and instructions of LLMs, calling for future work on automated self-evolving frameworks without human intervention.
    \item \textit{Decision-Maker}: LLMs generate direct actions or indirect advice for the agent to improve exploration efficiency. Huge computational overhead is a major issue in online interaction. Cost-effective methods to reduce the huge computational overhead of LLM in online RL is an important direction.
    \item \textit{Generator}: With the generative capability and world knowledge, LLMs are used as 1) a high-fidelity world model to reduce real-world learning cost; and 2) a language-based policy interpreter to explain the agent policy. Leveraging human instructions to improve the accuracy and generalizability of the world model and policy interpreter would be a crucial direction.
\end{itemize}


\bibliographystyle{IEEEtran}
\bibliography{references}

\end{document}